\documentclass{article}

\usepackage{PRIMEarxiv}

\usepackage[utf8]{inputenc} 
\usepackage[T1]{fontenc}    
\usepackage{hyperref}       
\usepackage{url}            
\usepackage{booktabs}       
\usepackage{amsfonts}       
\usepackage{nicefrac}       
\usepackage{microtype}      
\usepackage{lipsum}
\usepackage{graphicx}
\usepackage{multirow}
\usepackage{wrapfig}
\usepackage{algorithm}
\usepackage{subcaption}
\usepackage{amsmath}
\usepackage{algorithmic}
\usepackage{xcolor}
\newcommand{\lora}{LoRA}
\newcommand{\mlora}{MultiLoRA}
\newcommand{\llama}{LLaMA}
\newcommand{\llamas}{LLaMA-7B}
\newcommand{\ft}{fine-tuning}
\newcommand{\fpft}{full parameter fine-tuning}

\newcommand{\dft}{\Delta W^{FT}}
\newcommand{\dw}{\Delta W}
\newcommand{\dlora}{\Delta W^{\lora}}
\newcommand{\dmlora}{\Delta W^{\mlora}}

\newcommand{\lorac}[1]{\lora$_{r=#1}$}
\newcommand{\mcite}[1]{\cite{#1}}
\newcommand{\mlorac}[2]{\mlora$_{r=#1}^{n=#2}$}

\title{MultiLoRA: Democratizing LoRA for Better Multi-Task Learning
}

\author{
  Yiming Wang, Yu Lin, Xiaodong Zeng, Guannan Zhang \\
  Ant Group \\
  Shanghai, China\\
}

\begin{document}
\maketitle

\begin{abstract}
  \lora~achieves remarkable resource efficiency and comparable performance when adapting LLMs for specific tasks.
Since ChatGPT demonstrated superior performance on various tasks, there has been a growing desire to adapt one model for all tasks.
However, the explicit low-rank of \lora~limits the adaptation performance in complex multi-task scenarios.
\lora~is dominated by a small number of top singular vectors while \ft~decomposes into a  set of less important unitary transforms.
In this paper, we propose \mlora{} for better multi-task adaptation by reducing the dominance of top singular vectors observed in LoRA.
\mlora~scales \lora{} modules horizontally and change parameter initialization of adaptation matrices to reduce parameter dependency, thus yields more balanced unitary subspaces.
We unprecedentedly construct specialized training data by mixing datasets of instruction follow, natural language understanding, world knowledge, to cover semantically and syntactically different samples.
With only 2.5\% of additional parameters, \mlora{} outperforms single \lora~counterparts and \ft~on multiple benchmarks and model scales. Further investigation into weight update matrices of \mlora~exhibits reduced dependency on top singular vectors and more democratic unitary transform contributions\footnote{Our code is coming to GitHub soon.}.

\end{abstract}


\section{Introduction}

In recent years, Large Language Models (LLMs) have manifested unprecedentedly superior performance in various natural language processing tasks\mcite{bert,gpt3,roberta,llama}.
As model scales up, high-level multi-task capabilities\mcite{merge} emerges from LLMs.
Capabilities such as real-world knowledge, logic reasoning and arithmetic skills from one LLM proves the feasibility of "one model for all tasks".
However, scaling up LLMs by adding billions of parameters not only bring emergent abilities and grokking, but also dramatically increase training and down-stream adaptation costs.
For instance, parameter counts of \llama\mcite{llama}~series range from 7 billion to 65 billion, and GPT-3\mcite{gpt3} contains up to 175 billion parameters.
Full parameter fine-tuning these models for down-stream adaptation yields huge amount of memory footprint and thus requires prohibitively expensive hardwares.

To address the issue of hardware requirements for LLM adaptation, a solution called Parameter Efficient Fine-Tuning (PEFT) has been proposed.
PEFT methods reduce VRAM usage of cached optimizer states\mcite{lora} by only optimizing a fraction of model parameters while keeping the rest frozen.
Various PEFT methods, such as adapter\mcite{adapter}, p-tuning\mcite{ptuning2}, IA$^3$\mcite{ia3} and \lora\mcite{lora}, have been suggested.
Compared to other PEFT methods, \lora~possesses the advantages of: 1) high modularity for distribution, 2) mergeable weights for zero inference overhead.
While \lora~has proven successful in single-task adaptation, its performance in more intricate multi-task settings of generative AI remains unexplored.
Thus, a crucial question lingers: Can \lora~effectively adapts LLMs to complex multi-task scenarios as full parameter fine-tuning does?

Works on applying PEFT methods on multi-task learning scenarios are in literature, albeit with certain limitations\mcite{peftmt2,peftmt3,mt1,adamix}.
These proposed methods manage to improve multi-task benchmark performances with task information sharing or activation routing\mcite{adamix,peftmt2,peftmt3}.
However, these dedicated modules add unaffordable overhead to transformer inference, which hinders their industrial application\mcite{lora,ds}.
Another limitation is that the prior works focused on Natural Language Understanding (NLU), which may not be suitable for current generative LLMs. A mixture of NLU tasks are commonly used\mcite{peftmt2,peftmt3} despite that data samples of these tasks do not present much semantical or syntactical difference among them. More benchmarks on tasks of interest of generative LLMs, like instruction following, logic reasoning should be taken into consideration.

Therefore, the research goal of this paper is to adapt \lora~for better multi-task learning while maintaining modularity and zero inference overhead of \lora.
We firstly reveal the fundamental difference between \lora~and \fpft~with the help of Singular Value Decomposition (SVD, Section \ref{sec-loravsft}).
We found dominance of top singular vectors in \lora~while \ft~is more democratic, as the residual weight decomposes into larger sets of unitary transforms of smaller importance.
In order to mitigate the observed dominance, we propose to horizontally scale \lora modules  (Section \ref{sec-method-design}). \mlora~horizontally scales lora modules  to reduce parameter dependency. \mlora~divides \lora~along the rank, add learnable scaling factor and change the parameter initialization to enhance expressiveness of lora modules. Compared to conventional \lora, \mlora~produces more democratic weight update matrices as those of \fpft.

To better demonstrate the effectiveness of \mlora, we constructed a comprehensive dataset composed of various tasks relevant to generative LLMs.
A series of datasets from different domains are selected including instruction following\mcite{alpaca}, world knowledge\mcite{mmlu}, arithmetic reasoning\mcite{gsm8k} and NLU\mcite{superglue}.
Both context and target of samples in aforementioned datasets exhibit strong semantical and syntactical differences, thus augmenting adaptation difficulty.
With our multi-task datasets, we conducted extensive empirical experiments on \llama~ranging from 7B to 65B.
On benchmarks of MMLU\mcite{mmlu} and SuperGLUE\mcite{superglue}, we found \mlora~consistently outperforms \lora~even under smaller parameter budget and can perform on-par with full-parameter fine-tuning.
We further dive into obtained weight update matrices with SVD.
Side by side comparison to \fpft~suggests that \mlora~exhibit a higher degree of subspace overlapping and more similar singular value distribution, indicating successful democratization of unitary transforms contribution.

Therefore, our main contributions can be summarized as follows:
\begin{itemize}
   \item We find dominance of unitary transforms in weight update matrices of \lora, while \ft~produces more democratic contribution distribution.
   \item We propose MultiLoRA to mitigate dominance seen in \lora and democratize contributions of its unitary transforms.
   \item We propose a multi-task learning scheme based on mixture of tasks of interest of generative LLMs, to cover semantically and syntactically different samples. Our proposed \mlora~exhibits stronger consistency than \lora~and can outperform  \fpft~on various tasks and model scales.
\end{itemize}

\section{Related Work}
\subsection{PEFT}
PEFT methods lowers hardware requirement of model fine-tuning by significantly reducing trainable parameters and consequently optimizer states cached in VRAM.
By exploiting the local optimum of a pretrained model, a much smaller solution space brought by reduce trainable parameters helps PEFT methods achieve comparable tuning performance\mcite{delta1,delta2}.
PEFT can be classified into two categories: 1) reparameterization-based methods\mcite{bitfit, rome} that retrain a portion of the parameters and 2) addition-based methods that train additional parameters\mcite{lora,adalora,adapter}.
Recent works in PEFT focus on resource efficiency\mcite{adapter,ia3,lora,adalora}.
\lora\mcite{lora} fits incremental weights by decomposing them into low-rank matrices. (IA)$^3$ tunes hidden states with learned multipliers. AdaLoRA\mcite{adalora} adds importance-aware pruning mechanisms to further improve resource efficiency. There're also work focusing on ensemble learning with adapters. AdaMix\mcite{adamix} and UniPELT\mcite{unipelt} integrate existing PEFT methods into a unified framework to boost adaptation performance.

\subsection{Multi-Task Learning with PEFT}
In multi-task learning with PEFT, adapter is utilized for code summarization across different programming languages\cite{mt1}.
HyperFormer\mcite{peftmt2} assigns task-related weights to adapter\mcite{adapter} activations using shared hypernets  across layers and tasks.
Multitask Prompt Tuning\mcite{peftmt3} extends prompt tuning by firstly distilling from source prompts adapted for various tasks and further finetunes with low rank updates.
While these methods have shown effectiveness, the additional weights cannot be seamlessly integrated into the base model,  resulting in inevitable inference latency that is impractical for LLM serving\mcite{lora,ds}. Moreover, the emphasis in the multi-task setting has predominantly been on NLU tasks, disregarding the tasks that are of interest to generative LLMs.

\section{Method}

\subsection{Background}

Before formal explanation on design choices of \mlora, a few notations are proposed base on \llama~and \lora.

\subsubsection{LLaMA}
\llama~model consists of $L$ stacked decoder layers, where each
block contains two submodules: a multi-head attention (MHA) and a fully connected FFN.
Given the input sequence $\textbf{x}\in  \mathbb{R}^{n\times d}$, MHA performs the attention function in parallel h heads:
\begin{equation}
   \text{head}_i = \text{Softmax}(\frac{\textbf{x}W^{q\_proj}_{i} (\textbf{x}W^{k\_proj}_{i})^\top}{\sqrt{d_n}})\textbf{x}W^{v\_proj}_{i},\\
   \text{MHA}(\textbf{x})=\text{Concat}(\text{head}_1,\dots, \text{head}_n)W^{o\_proj},
\end{equation}
where $W^{q\_proj}_i,W^{k\_proj}_i,W^{v\_proj}_i \in \mathbb{R}^{d\times d_h}$ are query, key and value projections of  $i^{th}$ attention head and $W^{o\_proj}\in \mathbb{R}^{d\times d}$ is the output projection to aggregate multi-head outputs. $d_h$ is typically set to $d/$h.
The other important module is a MLP which consists of three linear transformations, namely \textit{up\_proj, down\_proj, gate\_proj} with a SwiGLU activation in between:
\begin{equation}
   \text{MLP}(\textbf{x}) = \text{SwiGLU}(\textbf{x}W^{up\_proj}(\textbf{x}W^{gate\_proj}))W^{down\_proj},
\end{equation}
where $W^{up\_proj}, W^{gate\_proj} \in \mathbb{R}^{d\times d_{mid}}, d_{mid}>d$ and $W^{down\_proj} \in \mathbb{R}^{d_{mid}\times d}$.
Layer normalization is applied before and after the attention module.\mcite{llama}

\subsubsection{Low-Rank Adaptation}

Given target module with weight $W\in \mathbb{R}^{d\times k}$, \lora~inserts two sequential low rank matrices to fit the residual weights for adaptation. The forward computation of adapted module writes as follow:
\begin{equation}
   \textbf{y}'=\textbf{y}+\Delta \textbf{y}= W\textbf{x} + BA\textbf{x},
\end{equation}
where $A\in \mathbb{R}^{d\times r}, B\in \mathbb{R}^{r\times k}$ with $r\ll \min(d,k)$. Either $A$ or $B$ is initialized with zeroes and the other is initialized with Kaiming Uniform\mcite{kaiming-init} to force $\Delta\textbf{y}=0$ at the very beginning. Analysis on weight update matrices suggest that \lora~work by enhancing existing feature transforms in original model weight\mcite{lora}.

\subsection{Difference between \lora{}{} and \ft}
\label{sec-loravsft}

\begin{figure}[htbp]
   \centering
   \begin{subfigure}[b]{.48\textwidth}
      \includegraphics[width=\textwidth]{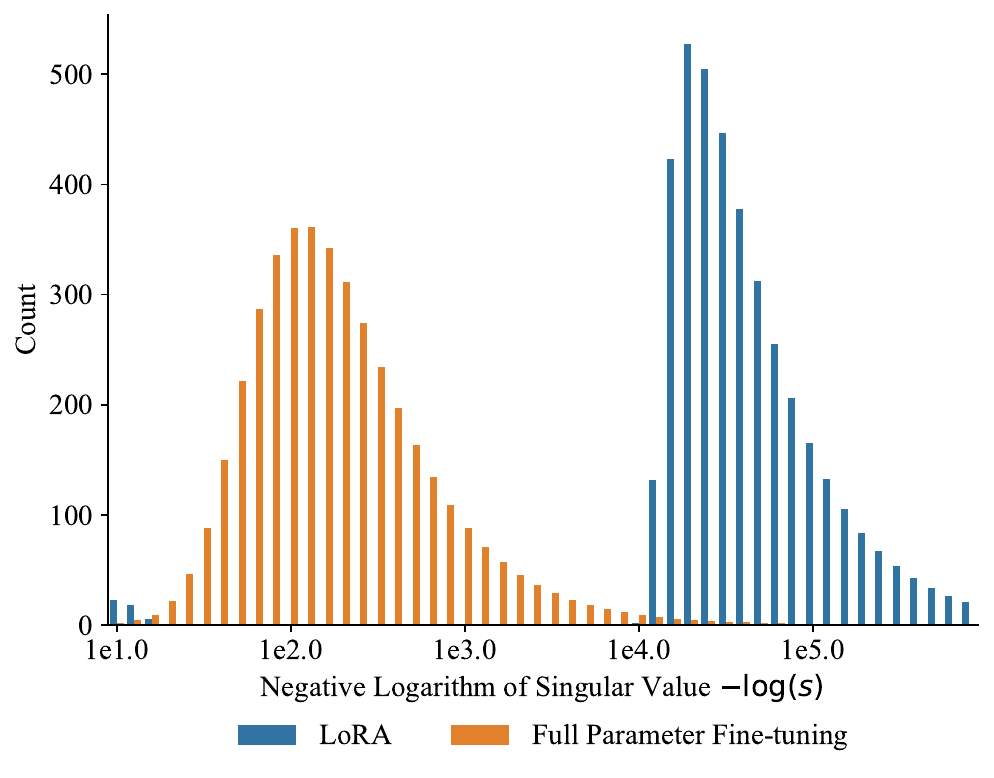}
      \caption{}
   \end{subfigure}
   \begin{subfigure}[b]{.48\textwidth}
      \includegraphics[width=\textwidth]{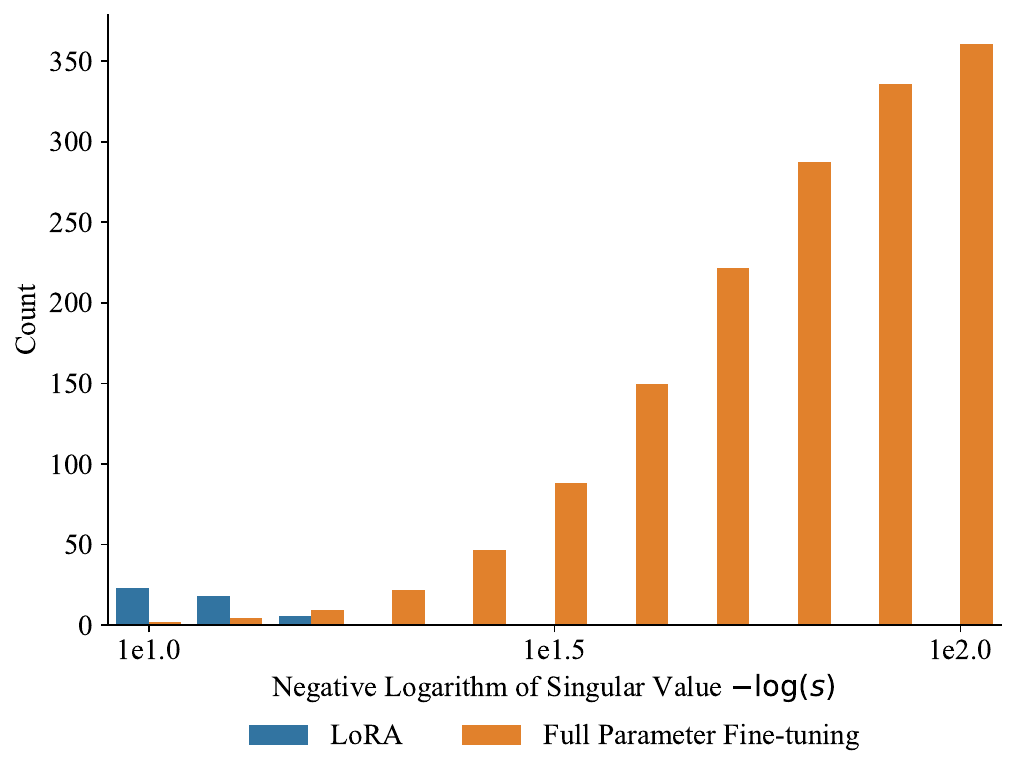}
      \caption{}
   \end{subfigure}

   \caption{Top singular value distribution of weight update matrix of $\Delta W_{v\_proj}$. (a) Complete view of the histogram. (b) Close-up view on top singular values.
      Both histograms are plotted based on on the negative logarithms of the singular values $-\log(s)$, where left end of horizontal axis represents larger singular values.
      Bell-shape curved of \fpft~indicates a democratic composition of a large number of relatively less important unitary transforms.
      On the hand, \lora{} heavily relies on a small group of important unitary transforms, which could hurt complex multi-task adaptation.
   }
   \label{fig-svd-diff}
\end{figure}

Although \lora{} achieves comparable performances to \ft~on many benchmarks, it is essential to understand the underlying differences between the two approaches. To shed light on this question, We train \llamas{} on Alpaca and MMLU using both methods\footnote{\lora~hyperparameters set to $r=64$ and $\alpha=64$} to get weight update matrices $\Delta W$ and conduct an analysis of the weight update matrices  using SVD.

Figure \ref{fig-svd-diff} illustrates singular value distribution of $\dft$ and $\dlora$.
For better visualization, we plot the negative logarithms of the singular values ($-\log(s)$) .
The empirical distribution of \ft~exhibits a bell-shaped curve while the distribution for \lora{} falls at both ends of the spectrum. The extreme bimodal distribution of \lora{} arises from the constraint that $Rank(\dlora)$ should not exceed $r$, resulting in at least ($k-r$) singular values being zero.

Interestingly, we also noticed an inverse trend in the counts of top singular values.
In LoRA, the count increased with the magnitude of the singular values, while fine-tuning exhibited the opposite behavior.
This suggests that LoRA predominantly relies on a small group of singular vectors, whereas fine-tuning distributes importance more evenly among singular vectors.
Such phenomenon can also be observed on \lora~trained on other datasets or publicly available \lora~weights, indicating observed dominance arises from the structural design of \lora~(refer to Appendix \ref{app-svd} for more examples).

Based on these findings, we can infer that the dominance observed in LoRA may limit its adaptation performance, particularly in complex multi-task scenarios that require enhancement of multiple distinct feature transforms.
$\Delta W$ of \fpft~decomposes into a larger set (more specifically equals rank of original weight matrix) of unitary transforms.
In contrast, LoRA's explicit rank limitation restricts it to decompose into a smaller number ($r$) of unitary transforms. As a result, the expressiveness of LoRA may be constrained compared to full parameter fine-tuning.

\subsection{Scaling \lora{}~to Democratize Unitary Transform Contribution}

\label{sec-method-design}

\begin{wrapfigure}{r}{0.4\textwidth}
   \centering
   \includegraphics[width=.38\textwidth]{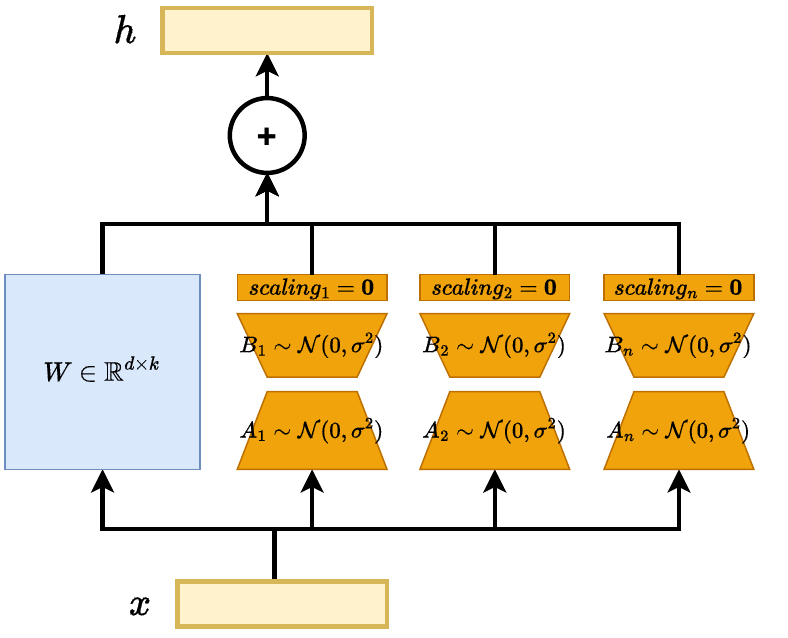}
   \caption{Overview of \mlora{}. Multiple parallel \lora~modules are used to adapt target weight matrix. Parameter initialization and zero-initialized scaling factor are introduced to democratize residual weight updates.
   }
   \label{fig-mlora-overview}
\end{wrapfigure}

In Section \ref{sec-loravsft}, we observe a small group of top singular vectors dominate weight update matrices of \lora{}~$\dlora$ while top singular vectors contribute more evenly to $\dft$. In order to match \ft~in complex task adaptation, we propose \mlora~aiming at producing less polarized weight update matrices $\Delta W$.
\mlora~inserts multiple parallel \lora{}s to reduce parameter sharing, changes parameter initialization to enable larger optimization search space and implement starting point initialization with a  learnable parameter.
Figure \ref{fig-mlora-overview} shows the overview of \mlora{}. There're 3 major difference to original \lora{}{}: horizontal scaling of \lora{}{} modules, scaling factors and parameter initialization.

\subsubsection{Scaling \lora{}{} Horizontally}
Given that \lora~performs closely despite scaling up the rank $r$, our key strategy of depolarization is parallelism.
Through parallelism, incremental activation $\Delta \textbf{y}$ is further decomposed into a series of independent variables, which allows for more degrees of freedom during optimization.
Bear in mind that under the same parameter budget, decomposing one large \lora~module into multiple small \lora s cannot augment rank of $\Delta W$ as $rank(AB)\leq \min(rank(A),rank(B))$ and $rank(A+B)\leq rank(A) + rank(B)$.
Corresponding weight matrices are noted as $\{A_i\in \mathbb{R}^{r\times k}\}_{i\in[1,n]},\{B_{i}\in \mathbb{R}^{d\times r}\}_{i\in[1,n]}$.
Thus, the forward computation of \mlora~writes as:
\begin{equation}
   \Delta \textbf{y}=\sum^n_{i=1} scaling_iB_iA_i\textbf{x},
\end{equation}
Comparing with the forward of \lora, \mlora~differs in that less parameter dependency brought to $\{B_i\}$.
Parallelism has no effect on $A$ as intermediate $m_i=A_i\textbf{x}$ is equivalent to reshape the result of $[A_0^\top,\dots ,A_n^\top]^\top\textbf{x}$. But here comes the major difference.

\subsubsection{Parameter Initialization}

Scaling \lora~horizontally allows for independent feature transform especially the up-projection of $\{B_i\}$. To further push the expressiveness of $\{B_i\}$, we change its parameter initialization to Kaiming-Uniform \mcite{kaiming-init} instead of all zeroes and consequently introduce a learnable scaling factor to implement the starting point initialization.

The zero initialization seen in $B$ of \lora~aims to keep activation unchanged before training. Such practice, we term as \textit{Starting Point Initialization}, is commonly seen in PEFT methods but may be implemented differently.
With starting point initialization, tuning a pretrained LLM essentially becomes optimizing in a much smaller parameter space around the local optimum of pretrained models.
However, zero initialization is a double-edged sword.
It is also infamous for introducing redundancy and breaks asymmetry\mcite{init1,kaiming-init}, yielding limited expressiveness of networks despite faster convergence speed during adaptation.

To take advantage of the starting point initialization and mitigate the drawbacks of zero initialization,
\mlora~changes initialization of $\{B_i\}$ to Kaiming-Uniform and implements stating point initialization with zero-initialized learnable scaling factors $scaling_i\in \mathbb{R}^k$.
Kaiming-Uniform has been shown to improve the generalization performance of neural networks and is the default parameter initialization method in PyTorch\mcite{torch} implementation.

\section{Experiments}
In this section, we evaluate our proposed method from three aspects, namely memory profile, throughput and downstream performance. All our experiments are conducted with LLaMA series\mcite{llama}, ranging from 7B to 65B.

\subsection{Experiment Setups}
\label{sec-exp-setup}

\begin{table}[htbp]
   \centering\begin{tabular}{llccccc}
      \textbf{Model Size}  & \textbf{Method}      & \textbf{MMLU} & \textbf{Boolq} & \textbf{MultiRC} & \textbf{RTE}  & \textbf{WIC}  \\ \hline
      \multirow{4}{*}{7B}  & Zero-Shot            & 35.1          & 66.5           & 42.3             & 57.0          & 49.4          \\
                           & FT                   & \textbf{45.3} & 87.6           & \textbf{84.5}    & \textbf{87.0} & \textbf{71.2} \\
                           & \lorac{96}           & 44.7          & 86.0           & 81.7             & 86.6          & 67.6          \\
                           & \mlorac{32}{3}(Ours) & 45.1          & \textbf{88.7}  & 83.8             & 85.6          & 70.2          \\
      \hline
      \multirow{4}{*}{13B} & Zero-Shot            & 46.9          & 65.0           & 43.4             & 60.6          & 49.5          \\
                           & FT                   & \textbf{51.3} & 87.1           & 85.7             & 90.8          & 74.3          \\
                           & \lorac{96}           & 51.0          & \textbf{87.3}  & \textbf{86.1}    & \textbf{91.7} & 69.9          \\
                           & \mlorac{32}{3}(Ours) & 51.3          & 86.7           & 84.7             & 91.4          & \textbf{75.4} \\
      \hline
      \multirow{4}{*}{30B} & Zero-Shot            & 57.8          & 74.6           & 46.9             & 53.4          & 50.0          \\
                           & FT                   & \textbf{59.2} & 89.3           & 87.9             & 92.8          & 74.0          \\
                           & \lorac{96}           & 58.8          & \textbf{89.7}  & 87.0             & 91.0          & 74.1          \\
                           & \mlorac{32}{3}(Ours) & 59.1          & 89.5           & \textbf{88.1}    & \textbf{93.1} & 74.1          \\
      \hline
      \multirow{4}{*}{65B} & Zero-Shot            & 63.5          & 73.6           & 48.3             & 59.6          & 51.3          \\
                           & FT                   & \textbf{64.6} & \textbf{91.6}  & 90.1             & \textbf{93.9} & 75.4          \\
                           & \lorac{96}           & 64.2          & 91.4           & 90.0             & 93.1          & 74.5          \\
                           & \mlorac{32}{3}(Ours) & 63.3          & 91.0           & \textbf{90.2}    & 93.5          & \textbf{74.7} \\\hline
   \end{tabular}
   \caption{Main results on MMLU and SuperGLUE using \llama{} of all scales trained in conventional single dataset setup.  MMLU is tested with 5-shot prompts and SuperGLUE are tested with zero-shot. \mlora, \lora~and \fpft produces similar results on single dataset setup.}
   \label{tab-exp-sd}
\end{table}
\subsubsection{Training Data}
\label{sec-exp-data}

To evaluate on tasks of interest of generative LLMs, we build multi-task datasets encompassing Alpaca\mcite{alpaca}  for instruction following, MMLU\mcite{mmlu} for world knowledge, GSM8K\mcite{gsm8k} for  arithmetic reasoning  and SuperGLUE\mcite{superglue} for NLU.
Therefore, our mixture of tasks covers semantically and structurally different samples.
In terms of source and target sequence length, samples from MMLU and SuperGLUE consist of single-choice questions with very short target lengths, typically one token. On the other hand, Alpaca and GSM8k contain longer target sequences.
From the aspect of task semantic, the subjects covered by each dataset differ. MMLU encompasses real-world knowledge across various domains such as humanities, STEM, and social sciences, offering different levels of difficulty. In contrast, Alpaca focuses primarily on aligning model output with human preferences. GSM8k train the models to generate logical and step-by-step responses to questions.

To ensure consistency in evaluation, we follow QLoRA\mcite{qlora}  and MeZO\mcite{mezo} to verbalize samples of MMLU and SuperGLUE, respectively. This verbalization process helps standardize input data across tasks, enabling fair comparisons.
During training, we introduce random shuffling to enhance the learning process and prevent any bias that may arise from the ordering of the samples.

\subsubsection{Baselines}
We use models from \llama{}\mcite{llama} series as the base model.
In our comparative analysis, we consider two baselines: full parameter fine-tuning (referred to as \textbf{FT}) and single \textbf{LoRA} (referred to as LoRA).
To establish a strong single LoRA baseline, we incorporate LoRA modules alongside all linear layers of LLaMA.
Specifically, we insert LoRA modules in
\textit{q\_proj, k\_proj, v\_proj, o\_proj, up\_proj, down\_proj, gate\_proj} modules in \llama. The more layers that are adapted by \lora{}, the better down-stream task performances will be\mcite{qlora,lora}.

For Boolq, MultiRC, RTE and WIC, we report zero-shot performances and we report 5-shot results for MMLU.
Instead of reporting the individual best task scores, we report the score of each task when the best average score is achieved to emphasize multi-task capability. The hyperparameter settings employed in our experiments are detailed in Appendix \ref{app_hparam}. All experiments are conducted using 8 A100 80G GPUs .
Python library PEFT\mcite{peft} is used to help implement \mlora~and \lora.
We use Deepspeed ZeRO-3\mcite{zero} for distributed training and offload optimizer states and model parameters for larger training throughput.

\subsection{Evaluation Results}

\begin{table}[htbp]
   \centering\begin{tabular}{llrrrrrrr}
      Model Size           & Method               & \# Params & MMLU          & Boolq         & MultiRC       & RTE           & WIC           & AVG.          \\\hline
      \multirow{5}{*}{7B}  & FT                   & 100\%     & 49.5          & 88.4          & 87.2          & 85.2          & \textbf{74.0} & 76.9          \\
                           & \lorac{96}           & 3.6\%     & 47.7          & 88.2          & 85.4          & 83.4          & 71.6          & 75.2          \\
                           & \lorac{160}          & 5.9\%     & 50.2          & 87.7          & 85.3          & 83.3          & 70.1          & 75.3          \\
                           & \mlorac{32}{3}(Ours) & 3.6\%     & 51.2          & 87.8          & 88.7          & \textbf{89.7} & 70.8          & 77.6          \\
                           & \mlorac{32}{5}(Ours) & 6.0\%     & \textbf{51.4} & \textbf{88.5} & \textbf{89.4} & 89.4          & 71.4          & \textbf{78.0} \\\hline
      \multirow{5}{*}{13B} & FT                   & 100\%     & 51.4          & 89.2          & 89.3          & \textbf{91.3} & \textbf{75.1} & 79.2          \\
                           & \lorac{96}           & 2.9\%     & 49.7          & \textbf{89.7} & 88.5          & 87.0          & 71.5          & 77.2          \\
                           & \lorac{160}          & 4.8\%     & 50.4          & 89.4          & 88.4          & 87.6          & 72.1          & 77.5          \\
                           & \mlorac{32}{3}(Ours) & 2.9\%     & 52.6          & 89.4          & \textbf{89.9} & 86.9          & 74.1          & 78.5          \\
                           & \mlorac{32}{5}(Ours) & 4.8\%     & \textbf{52.9} & 89.3          & 89.5          & 90.3          & 74.3          & \textbf{79.4} \\\hline
      \multirow{5}{*}{30B} & FT                   & 100\%     & 57.5          & 90.5          & \textbf{91.0} & 91.7          & \textbf{75.9} & \textbf{81.3} \\
                           & \lorac{96}           & 2.2\%     & 57.1          & 90.2          & 90.5          & 90.5          & 74.0          & 80.4          \\
                           & \lorac{160}          & 3.7\%     & 56.8          & 90.8          & 90.1          & 89.9          & 73.8          & 80.2          \\
                           & \mlorac{32}{3}(Ours) & 2.3\%     & \textbf{58.4} & 90.6          & 90.5          & 91.5          & 74.9          & 81.1          \\
                           & \mlorac{32}{5}(Ours) & 3.8\%     & 58.0          & \textbf{91.7} & 90.6          & \textbf{91.9} & 75.2          & 81.2          \\\hline
      \multirow{5}{*}{65B} & FT                   & 100\%     & \textbf{66.4} & 91.7          & \textbf{91.3} & \textbf{93.9} & 76.5          & \textbf{83.9} \\
                           & \lorac{96}           & 1.8\%     & 65.9          & 91.3          & 90.8          & 92.4          & 75.1          & 83.1          \\
                           & \lorac{160}          & 3.1\%     & 65.8          & 90.9          & 90.4          & 93.6          & 75.5          & 83.2          \\
                           & \mlorac{32}{3}(Ours) & 1.8\%     & 65.9          & 91.5          & 90.5          & 93.8          & 76.2          & 83.5          \\
                           & \mlorac{32}{5}(Ours) & 3.1\%     & 66.3          & \textbf{91.8} & 90.1          & 93.3          & \textbf{76.6} & 83.6          \\
      \hline
   \end{tabular}
   \caption{
      Evaluation results on MMLU and SuperGLUE using \llama{} of all scales trained on our mixture. We report score of each task when the best average score is achieved throughout training. MMLU is tested with 5-shot prompts and SuperGLUE are tested with zero-shot. \mlora~produces better and more consistent results compared to \lora.}
   \label{tab-exp-md}
\end{table}

Benchmark performances trained on mixed data are listed in Table \ref{tab-exp-md}.
Comparing these results to those obtained from single datasets (listed in Table \ref{tab-exp-sd}), training on mixed datasets generally leads to benefits across all tasks and model scales, resulting in improved benchmark scores to varying extents.
Based on these findings, we draw the following conclusions:.

\textbf{\mlora~consistently outperforms \lora~and achieves better results than full parameter fine-tuning on smaller models.}
Across all benchmarks and model scales, \mlora~demonstrates stronger data fitting capabilities and outperforms the \lora~counterpart with the same parameter budget by a notable margin. For instance, \mlora~improves upon \lora~'s performance by 3.5\% on MMLU for LLaMA-7B and by 5.9\% on RTE for the same model. On average, \mlora~surpasses \lora~in terms of the evaluated tasks' average score by 2.8\%. Notably, \mlora~even outperforms full parameter fine-tuning on smaller models (7B and 13B), only slightly falling behind on larger scales. Specifically, \mlora~achieves an average score improvement of 1.1\% compared to full parameter fine-tuning on LLaMA-7B, and a slight 0.3\% decrease on LLaMA-65B. These significant improvements highlight \mlora~'s superior capability in complex multi-task adaptation.

\textbf{\mlora{} exhibits small performance fluctuations comparable to \fpft~in complex multi-task learning scenarios.}
On smaller models, \lora~tends to show performance variability, with more frequent fluctuations between different tasks. For example, on LLaMA-7B, compared to \fpft~, MultiRC, RTE, and WIC scores exhibit fluctuations of over 3\% in \lora~. In contrast, both \fpft~and MultiLoRA yield consistent individual task scores. The observed fluctuations in \lora~can be attributed to the dominance of top singular vectors, as noted in Section \ref{sec-loravsft}, where a small number of unitary transforms carry significant importance.

\textbf{In the single dataset setting, \mlora~performs similarity to \fpft~and \lora.} Table \ref{tab-exp-sd} shows evaluation results of models trained on single dataset. Across the 5 tested tasks and 4 scales, \fpft~performs the best on 9 combinations, while MultiLoRA and LoRA perform the best in 7 and 5 combinations, respectively (MultiLoRA performs equally to LoRA on WIC for LLaMA-30B). Based on these observations, we cannot definitively declare one approach as superior. However, in the multi-task setting, \mlora~and \fpft~demonstrate better performances compared to \lora.

\subsection{Resources \& Throughput Analysis}

Training throughput, VRAM usage and inference latency are crucial for generative LLMs.
In this section, In this section, we thoroughly examine the resource usage and throughput of \mlora~as we scale up the number of parallel \lora~modules $n$.
We primarily focus on VRAM usage and throughput during training as \mlora~inherits zero inference overhead from original \lora.
Our benchmark protocol involves training \llamas{} on sequences of 1024 tokens using 8 A100 GPUs and recording the peak VRAM usage and throughput\footnote{We train \mlora~with individual rank of 32.}.  Deepspeed ZeRO-3 and model parameter offload are activated to better evaluate impacts brought by \mlora.
\begin{figure}[htbp]
   \centering
   \begin{subfigure}[b]{.45\textwidth}
      \includegraphics[width=\textwidth]{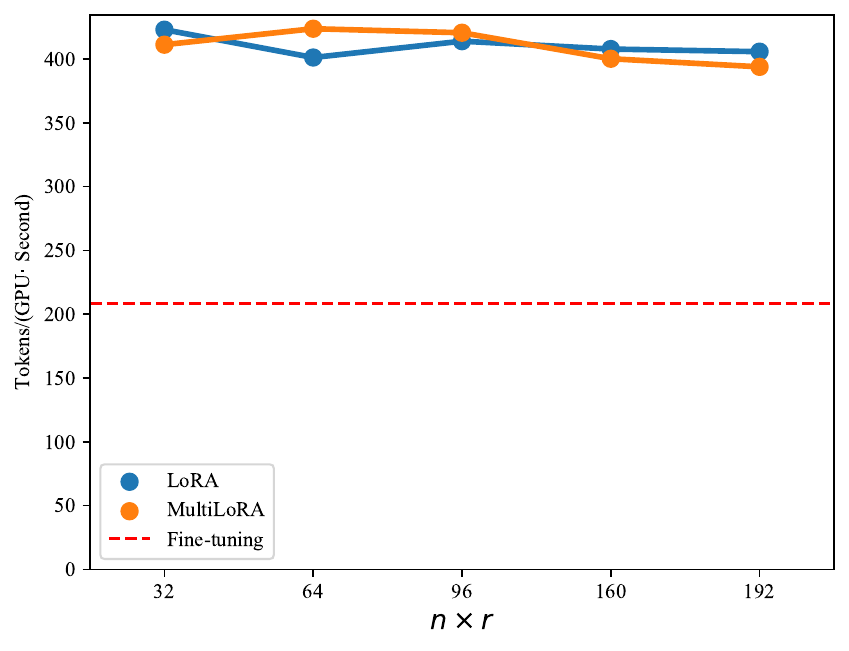}
      \caption{Throughput}
   \end{subfigure}
   \begin{subfigure}[b]{.45\textwidth}
      \includegraphics[width=\textwidth]{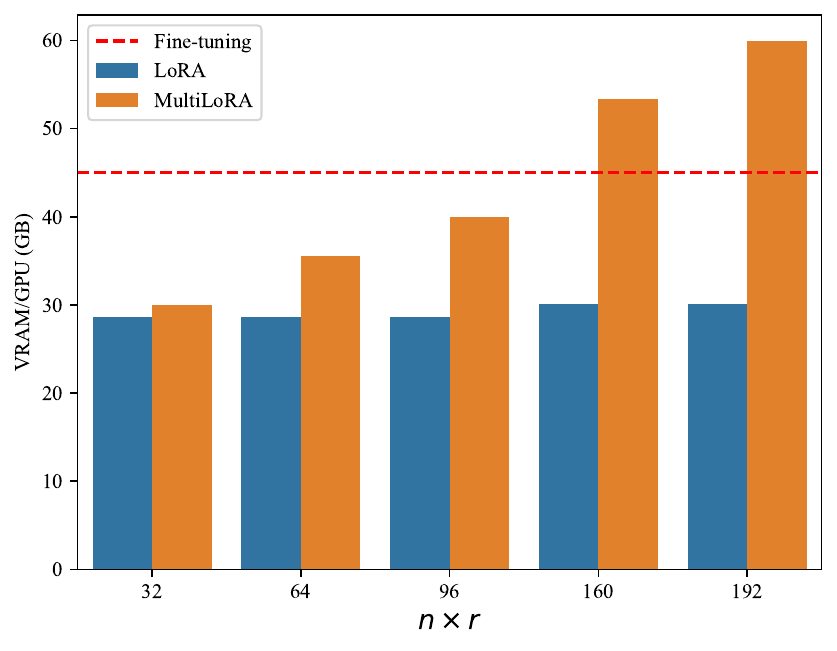}
      \caption{VRAM}
   \end{subfigure}
   \caption{(a) Throughput and (b) peak VRAM usage benchmarked when training \llamas with sequences of 1024 tokens and batch size of 1. $n\times r$ on horizontal axis indicates total rank of \lora~and \mlora. Thanks to high parallelism of \mlora, training throughput is almost identical to \lora. VRAM usage scales up linearly with the number of parallel \lora~modules.}
   \label{fig-throughput_vram}
\end{figure}

Results are listed in Figure \ref{fig-throughput_vram}.
On the horizontal axis, $n\times r$ denotes the equal total rank of \mlora{} ($n$ parallel \lora~of rank $r$) and \lora{} (one \lora~of rank $n\times r$).

Training throughput is one of the advantages of using \mlora~as other PEFT methods.
A limitation with \fpft{} lies in the fact that cached optimizer states can consume a significant portion of the VRAM.
Specifically, when training a 7B model with AdamW\mcite{adamw}, cached optimizer states can occupy up to 70\% of the available VRAM, and over 48\% with SGD\mcite{sgd}.
Thanks to the significantly reduced number of trainable parameters, finite VRAM can be leveraged to load more data samples, leading to larger training throughput.
Additionally, due to the parallelism inherent in \mlora{}, multiple \lora~modules do not introduce notable latency and the throughput remains close to that of \lora, around 400 tokens per GPU per second.
In our benchmarking, the throughput of \mlora{} is almost twice that of \fpft{} (208 tokens per GPU per second).

For VRAM usage, peak memory scales up much faster than \lora.
In order to optimize multiple parallel \lora~modules, multiple copies of activations should be cached in VRAM. Therefore, one major drawback of \mlora~is activation VRAM usage scales linearly with number of parallel \lora~modules which can be unaffordable in long sequence training.
In our benchmark, training \llamas~with sequences of 1024 tokens with $n=5$ would use more VRAM than \fpft.

\section{Understanding \mlora}

In this section, we apply SVD on weight update matrices trained with \llamas~in Section \ref{sec-exp-setup} to investigate why \mlora~outperforms \lora{} in complex task adaptation. Specifically, subspace similarity and magnitudes of singular value are thoroughly studied for \mlora, \lora{}~and \ft.

\subsection{Comparison with Fine-tuning}
\label{sec-under-vsft}
to demonstrate a higher degree of similarity to full parameter fine-tuning of \mlora, we utilize SVD to compare weight update matrices $\dw$ of \lora{}~and \mlora. Specifically, we focus on comparing the subspace coverage of singular vectors and the magnitudes of singular values.

As for subspace similarity of singular vectors, we follow \mcite{lora} to use $\phi(\dw',\dw,i,j)$ in Equation \ref{eq-simi}, the Frobenius norm of cosine similarity between top-i and top-j singular vectors of two weight update matrices.
\begin{equation}
   \label{eq-simi}
   \phi(\dw',\dw,i,j)=\frac{\|U_i^\top U'_j \|_F^2}{\min(i,j)}\in [0,1],
\end{equation}
where $U_i=U[:,:i]$ and $U'_j=U'[:,:j]$ are stacked top-i and top-j singular vectors.

Moreover, the magnitudes of singular values offer valuable insights into the relative importance of each singular vector. Larger singular values signify a greater contribution to the overall data representation. In Section \ref{sec-loravsft}, we find $\dlora$~is very polarized as proportion of top singular values is largest.
By comparing the singular value distribution, we want to find out whether \mlora~manages to balance contribution of  each  singular vectors.

\subsubsection{Subspace Comparison}
\label{sec-subspace}
\begin{figure}[htbp]
   \centering
   \includegraphics[width=\textwidth]{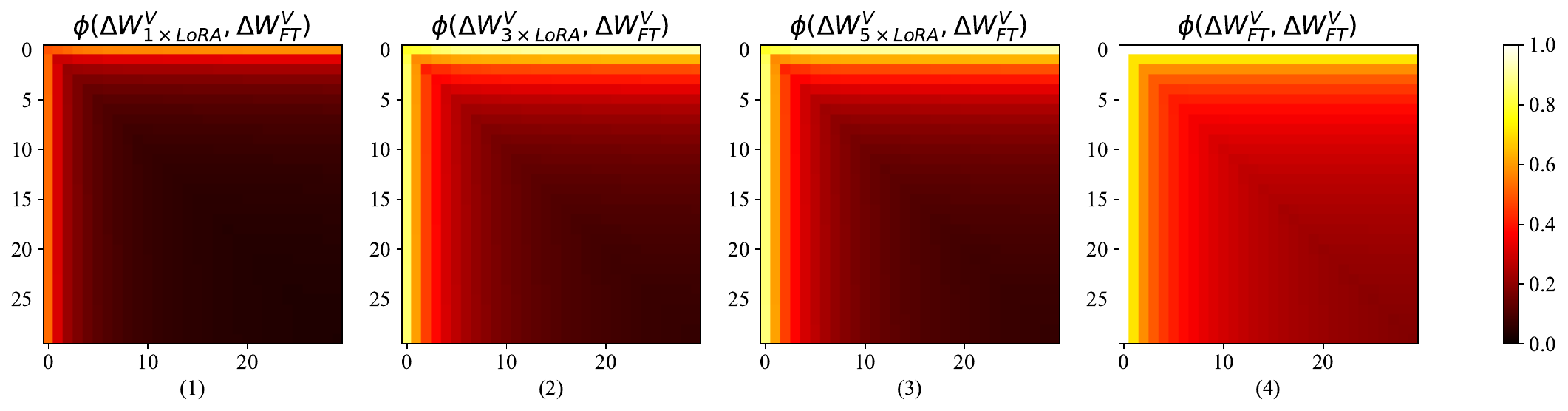}
   \caption{Subspace similarity to \ft~of \lora{}~(\textbf{1}), \mlora~(\textbf{2, 3}) and \ft~with a different random seed (\textbf{4}). \lora{}~(\textbf{2}) and \mlora~(\textbf{3}) share same parameter budget but \mlora~exhibits stronger subspace similarity to \ft. Heatmap of \mlorac{32}{3} does not differ much from that of \mlorac{32}{5}. Only $i,j\in[1,30]$ are presented for better visibility.}
   \label{fig_loravsft_simi}
\end{figure}
\label{sec-simi_com}
Orthonormal singular vectors define the "direction" of data transform. By measuring the subspace overlapping with $\phi(\dw',\dw)$, we can measure the degree of similarity between two transforms.
We randomly choose value projection of the 15$th$ decoder layer to calculate $\phi(\dlora,\dft)$ and $\phi(\dmlora,\dft)$. Similarity between \ft~of two different runs $\phi(\Delta W^{FT'},\dft)$ is also calculated for reference.

\textbf{\mlora~resembles \ft~more than \lora~in terms of subspace span.}
According to visualization in Figure \ref{fig_loravsft_simi}, \mlorac{32}{3}~exhibits stronger resemblance to \ft~than \lorac{96} under the same parameter budget, indicating that subspace of the weight update matrix of \mlora~is closer to that of \ft.
Heatmap of \lora{}~is generally dimmer but top singular vectors still present overlapping of subspaces to \ft~to some degree.

\textbf{Scaling up $n$ does not necessarily augment \mlora~subspace similarity to \ft.} Another thing to behold is barely visible difference between \mlorac{32}{3} and \mlorac{32}{5}, meaning that increasing parallel \lora{}~number $n$ does not necessarily make subspace closer to \ft.
The same trend can be observed on other weights of different depths in decoder stack (more at Appendix \ref{app-simi}).

\subsubsection{Singular Value Distribution Comparison}
\label{sec-sv-com}
\begin{figure}[htbp]
   \centering
   \includegraphics[width=.9\textwidth]{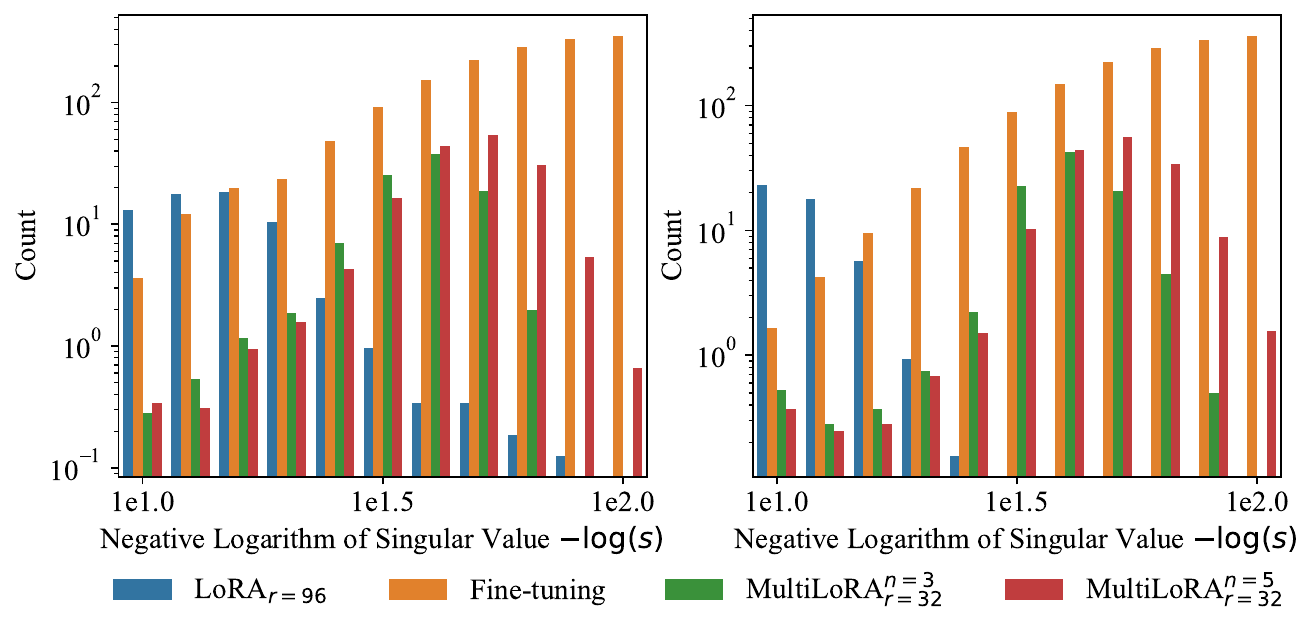}
   \caption{Singular value distribution of weight update matrices $\Delta W$ of \textit{k\_proj} (\textbf{Left}) and \textit{v\_proj} (\textbf{Right}). Our proposed \mlora~exhibits higher degree of resemblance to \ft. Scaling up $n$ produces more democratic unitary transform contributions.}
   \label{fig-sv-v}
\end{figure}
In previous Section \ref{sec-simi_com}, we measure the subspace similarity between unitary singular vectors but without knowing the importance of aforementioned singular vectors, we cannot conclude the higher resemblance between \mlora~and \ft.
Thus, we investigate into singular value distribution by plotting histogram of singular value as in Section \ref{sec-loravsft}.

For $\Sigma=diag(s)$ obtained from $\text{SVD}(\dw)$, we count the number of $s$ over a series of thresholds and average the statistics of the same module over different depths of decoder layers.
We calculate negative logarithms $-\log(s)$ for better visibility since more than 95\% of singular values are within $[0,1]$.
Results are shown in Figure \ref{fig-sv-v}.

\textbf{\mlora~balances subspace contributions compared to \lora.} \mlora{} shows similar distribution as \ft~ where number of singular value of \mlora~decreases with its magnitude.
Given the explicit low rank $r<<d$, \lora{}~shows heavy reliance on a small group of top singular vectors but \mlora democratizes contributions of singular vectors.

\textbf{Scaling up $n$ makes \mlora~amplify features at a more fine-grained level.} Comparing \mlorac{32}{3} and \mlorac{32}{5}, histograms of $\{s|-\log(s)>1e1.6\}$ are almost identical but \mlorac{32}{5} shows wider spectrum as proportion of small singular values increases.
A wider spectrum covering small singular values enables more fine-grained fitting of $\Delta W$ as \ft.

\subsection{Comparison among \mlora}

\begin{figure}[htbp]
   \centering
   \includegraphics[width=\textwidth]{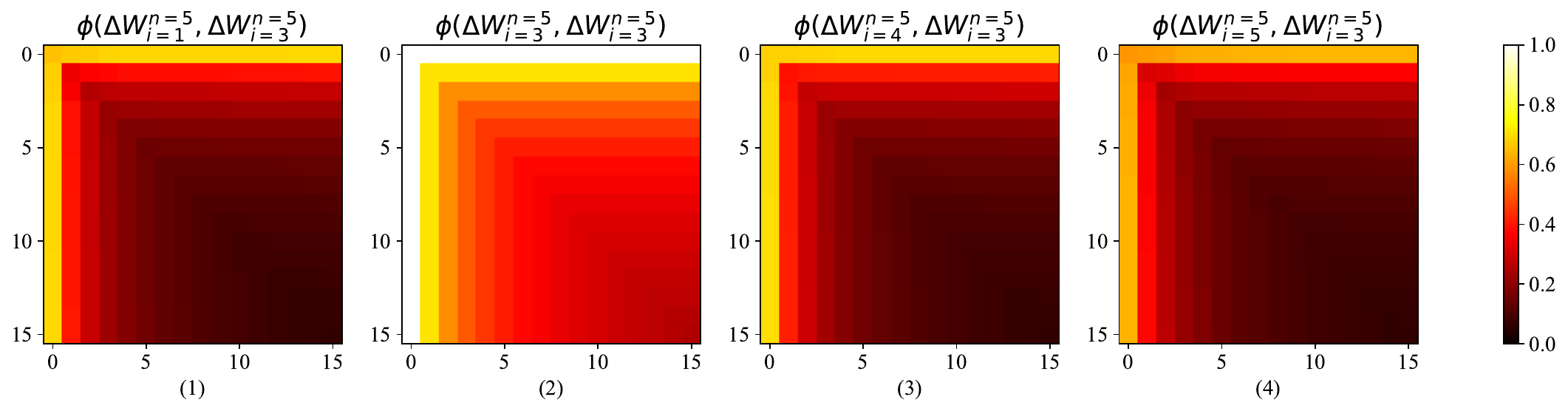}
   \caption{Subspace similarity between parallel module of \mlora. We analyze the \mlora~targeting \textit{down\_proj} in the first decoder layer. Each individual module produces close but not identical subspaces, thus augmenting the general expressiveness of \mlora. }
   \label{fig-intra}
\end{figure}
In this section, to demonstrate \mlora~accomplishes the design goal of depolarization, we compare in pair sub-\lora s.
From heatmap, subspace similarity between top-1 singular vectors is around 0.6.
Comparison between $\Delta W_{i=5}$ and $\Delta W_{i=3}$ shows relatively low similarity. The variance of subspace similarities indicates a more fine-grained pattern decomposition.

\subsection{Underlying Mechanisms of \lora~and \mlora}

In previous sections, we study the difference in subspace similarity and singular value distribution by applying SVD on weight update matrices of experimented methods.
Our observations shed light on underlying mechanisms of \lora~and \mlora. From the singular value distribution of \ft, we learn that \ft~fits residual weights by aggregating a large number (usually equals to rank of weight matrix) of relatively less important unitary transforms.
Given the low rank limitation, \lora~and \mlora~fits residual weights with $r\ll min(d,k)$ unitary transforms.
The low subspace similarity to \ft~and dominance in singular value distribution observed in \lora~show that \lora~tends to decompose the residual weights into unitary transforms of large importance.
Meanwhile, \mlora~democratizes influences of unitary transforms by assigning smaller importance to its unitary transforms similar to \ft. With democratized unitary subspaces, \mlora~produces better complex multi-task learning performance.

\section{Conclusion}
In conclusion, our study introduces MultiLoRA, a novel approach that enhances multi-task adaptation in language models.
By mitigating the dominance  of unitary transforms of LoRA, we successfully improve performance in complex multi-task scenarios.
Our proposed method focuses on scaling LoRA modules horizontally and modifying parameter initialization to reduce parameter dependency, thereby creating more balanced unitary subspaces.
Additionally, we construct a comprehensive dataset covering a wide range of tasks of interest for generative LLMs.
Through extensive experimentation, we have demonstrated that MultiLoRA outperforms single \lora~and achieves comparable performance to fine-tuning across multiple benchmarks and model scales. \mlora~stabilizes multi-task adaptation especially for smaller models.
Furthermore, our investigation into weight update matrices reveals a significant reduction in dependency on top singular vectors and a more equitable contribution of unitary subspaces in \mlora. Overall, MultiLoRA provides an efficient and effective solution for multi-task adaptation in language models.


\newpage
\bibliographystyle{unsrt}
\bibliography{references}

\newpage
\appendix
\section{Hyperparameters}
\label{app_hparam}
We list hyperparameters used in our experiments in the Table \ref{app-exp-hparam}.
Batch size of 32 is achieved for \llama-30B and \llama-65B with gradient accumulation. We use default values give by Huggingface transformers\mcite{transformers} trainer for most of the optimizer hyperparameters.
\begin{table}[htbp]
    \centering\begin{tabular}{lcc}
        \textbf{Expeiment}         & \textbf{Hyperparameters} & \textbf{Values} \\\hline
                                   & Batch Size per GPU       & 32              \\
                                   & Number of Epochs         & 2               \\
        \hline
        \multirow{4}{*}{Fine-Tune} & Learning Rate            & 5e-6            \\
                                   & LR Schedule              & Linear          \\
                                   & Optimizer                & AdamW           \\
                                   & Warmup Ratio             & 0.05            \\
        \hline
        \multirow{4}{*}{\lora{}}   & Learning Rate            & 5e-5            \\
                                   & LR Schedule              & Linear          \\
                                   & Optimizer                & AdamW           \\
                                   & Warmup Ratio             & 0.05            \\
        \hline
        \multirow{4}{*}{\mlora}    & Learning Rate            & 5e-5            \\
                                   & LR Schedule              & Linear          \\
                                   & Optimizer                & AdamW           \\
                                   & Warmup Ratio             & 0.05            \\
        \hline
    \end{tabular}
    \caption{Training Hyperparameters used in our experiments.}
    \label{app-exp-hparam}
\end{table}

\section{Singular Value Distribution}
\label{app-svd}

\begin{figure}[htbp]
    \centering
    \begin{subfigure}[b]{.48\textwidth}
        \includegraphics[width=\textwidth]{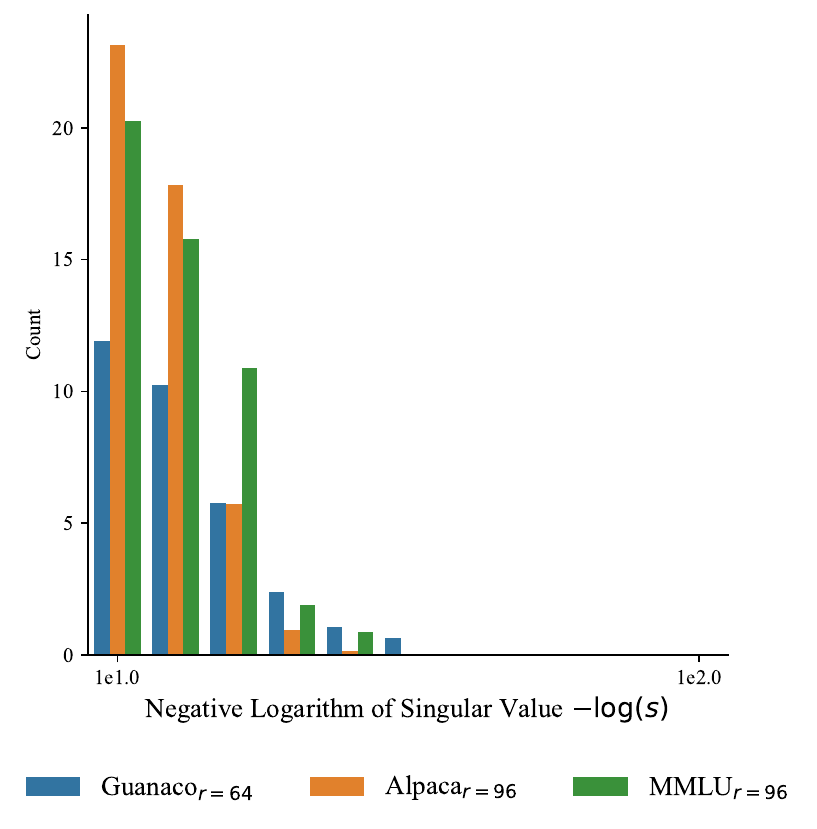}
        \caption{\textit{v\_proj}}
    \end{subfigure}
    \begin{subfigure}[b]{.48\textwidth}
        \includegraphics[width=\textwidth]{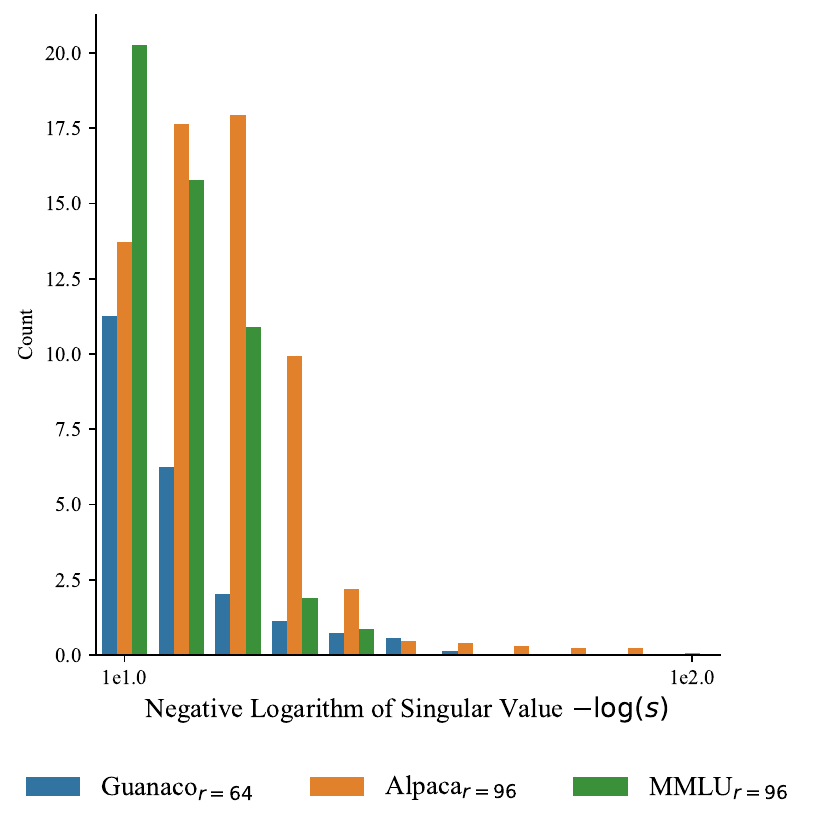}
        \caption{\textit{q\_proj}}
    \end{subfigure}
    \caption{Singular Value Distribution of (a) \textit{v\_proj} and (b) \textit{q\_proj} of weight update matrices trained on different datasets. }
    \label{fig-rank-app}
\end{figure}

In Section \ref{sec-loravsft}, we find that \lora's weight update matrices of ~are dominated by small group of unitary transforms.
To further support this, we analyzed LoRA modules obtained from training on various datasets or using publicly available resources.
We use \lora~modules obtained by training on public available datasets (MMLU and Alpaca) or downloading publicly available resources (Guanaco\footnote{Downloadable at \href{https://huggingface.co/timdettmers/guanaco-7b/tree/main}{https://huggingface.co/timdettmers/guanaco-7b/tree/main}}).

Figure \ref{fig-rank-app} plots histograms of singular values of weight update matrices of \textit{q\_proj} and \textit{v\_proj}.
To enhance visualization, the negative logarithm of the singular values $(-\log(s))$ is calculated, given that most values are smaller than 0.1

Mean values are used to aggregate statistics across all decoder layers.
The histograms for both modules exhibit a striking similarity. The triangular shape of the histograms indicates the dominance of the top singular vectors, as mentioned in Section  \ref*{sec-loravsft}. It is worth noting that this dominance arises from the inherent design of LoRA, as we do not deliberately alter its structure or use unconventional datasets.

\section{Subspace Similarities of other modules of different depth}
\label{app-simi}

In Section \ref{sec-simi_com}, we use cosine similarity between top singular vectors to measure subspace overlap of $\dw$. Here, we present more visualizations on different modules from different depths of the decoder stack.
\begin{figure}[htbp]
    \centering
    \begin{subfigure}[b]{\textwidth}
        \includegraphics[width=\textwidth]{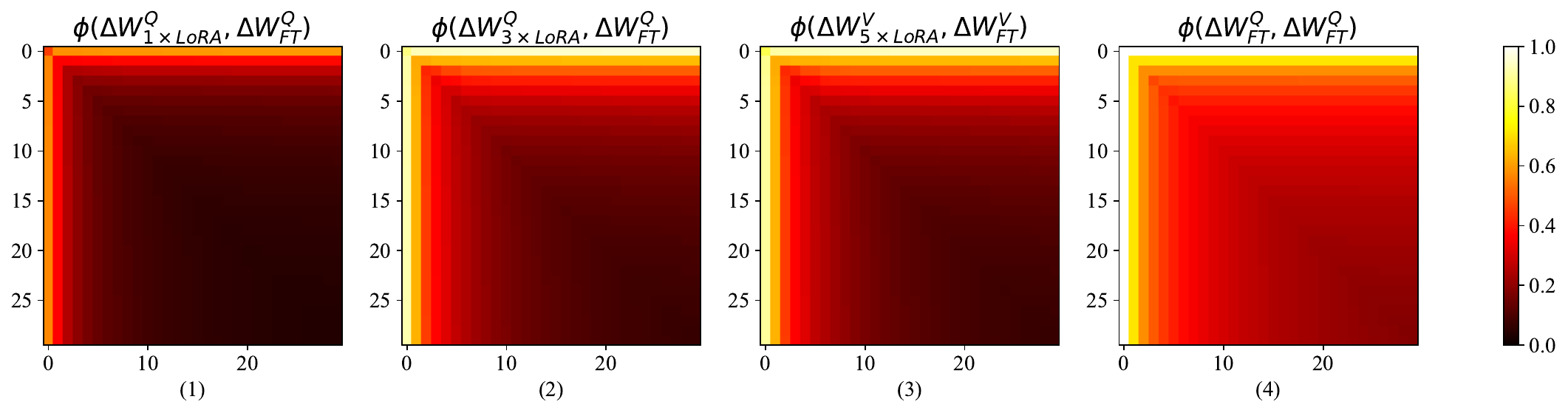}
        \caption{\textit{up\_proj} of MLP in the 28\textit{th} decoder layer.}
    \end{subfigure}
    \begin{subfigure}[b]{\textwidth}
        \includegraphics[width=\textwidth]{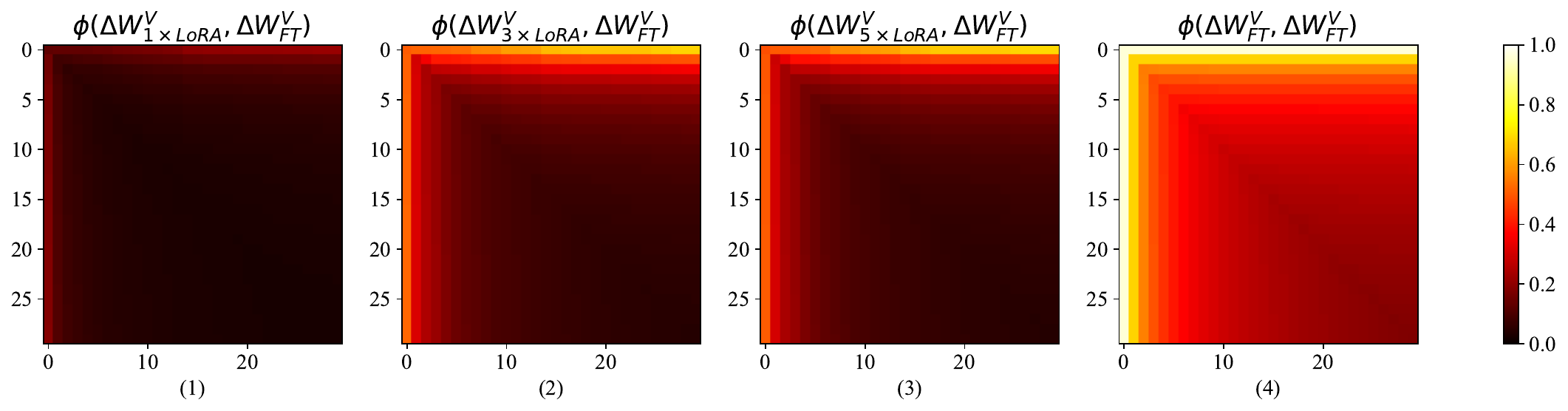}
        \caption{\textit{q\_proj} of self attention in the 1\textit{st} decoder layer.}
    \end{subfigure}
    \begin{subfigure}[b]{\textwidth}
        \includegraphics[width=\textwidth]{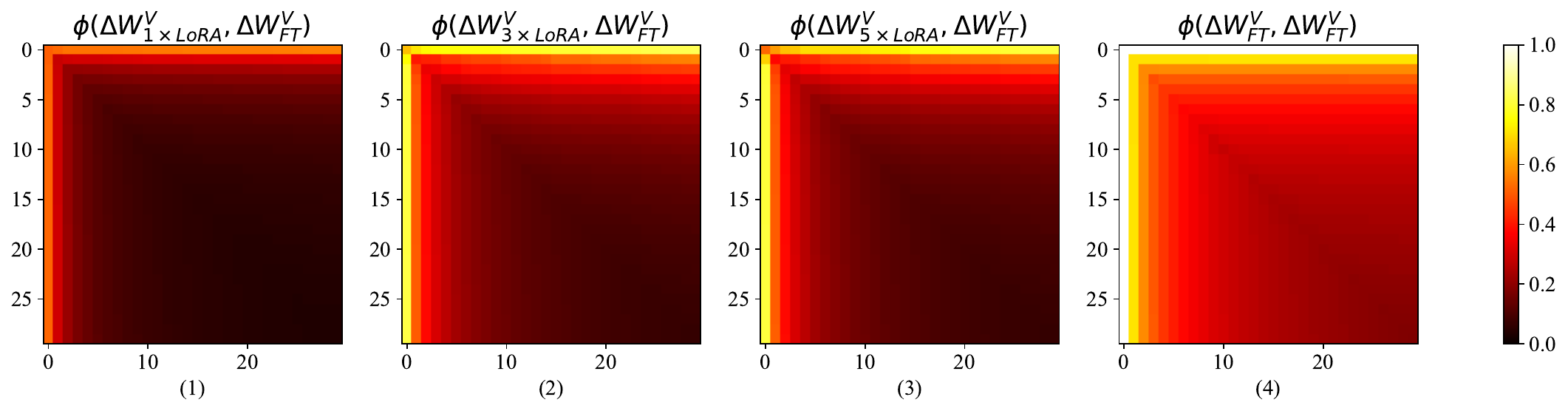}
        \caption{\textit{k\_proj} of self attention in the 14\textit{th} decoder layer.}
    \end{subfigure}
    \caption{Subspace similarity of \lora~and \mlora~to \ft of different modules at different depths.}
    \label{fig:subfigure}
\end{figure}

We randomly choose \textit{up\_proj}, \textit{q\_proj}  and \textit{k\_proj} of MLP and self attention module at different depths to compare weight update matrices of \lora~and \mlora~to \ft.
The heatmap visualization shows higher degree of similarity of \mlora~as observed in Section \ref{sec-subspace}.
Our observation sheds light on the mechanism of \lora that the residual weight is decomposed into a small number group of unitary transform of large importance. Important unitary transforms of small number hinders the model handling complicated multi-task learning. Meanwhile, \mlora~manages to fits the residual weight more similar to \ft~which gathers a larger number of relatively less important transforms.

\end{document}